\documentclass{article}

     \usepackage[preprint]{neurips_2018}

\usepackage[utf8]{inputenc} 
\usepackage[T1]{fontenc}    
\usepackage{hyperref}       
\usepackage{url}            
\usepackage{booktabs}       
\usepackage{amsfonts}       
\usepackage{nicefrac}       
\usepackage{microtype}      
\usepackage{makecell}
\usepackage{graphicx}
\usepackage{subfigure}
\usepackage{bbm}

\title{Pairwise Learning for Neural Link Prediction}

\author{%
  Zhitao Wang \\
  WeChat Pay, Tencent\\
  \texttt{zhitaowang@tencent.com} \\
   \And
   Yong Zhou \\
   WeChat Search, Tencent\\
   \texttt{joyceyzhou@tencent.com} \\
   \AND
   Litao Hong \\
   WeChat Pay, Tencent\\
   \texttt{brianlthong@tencent.com} \\
   \And
   Yuanhang Zou \\
   WeChat Search, Tencent\\
   \texttt{yuanhangzou@tencent.com} \\
   \And
   Hanjing Su \\
   WeChat Pay, Tencent\\
   \texttt{justinsu@tencent.com} \\
   \And
   Shouzhi Chen \\
   WeChat Pay, Tencent\\
   \texttt{easychen@tencent.com} \\
}

\begin{document}

\maketitle

\begin{abstract}
In this paper, we aim at providing an effective \textbf{P}airwise \textbf{L}earning for \textbf{N}eural \textbf{L}ink \textbf{P}rediction (PLNLP) framework. The framework treats link prediction as a pairwise learning to rank problem and consists of four main components, i.e., neighborhood encoder, link predictor, negative sampler and objective function. The framework is flexible that any generic graph neural convolutions or link prediction specific neural architectures could be employed as neighborhood encoder. For link predictor, we design different scoring functions, which could be selected based on different types of graphs. In negative sampler, we provide several sampling strategies, which are problem specific. As for objective function, we propose to use an effective ranking loss, which approximately maximizes the standard ranking metric AUC. We evaluate the proposed PLNLP framework on 4 link property prediction datasets of Open Graph Benchmark (OGB), including \texttt{ogbl-ddi}, \texttt{ogbl-collab}, \texttt{ogbl-ppa} and \texttt{ogbl-ciation2}. PLNLP achieves \textbf{top 1} performance on \texttt{ogbl-ddi} and \texttt{ogbl-collab}, and \textbf{top 2} performance on \texttt{ogbl-ciation2} only with basic neural architecture. The experimental results demonstrate the effectiveness of PLNLP.
\end{abstract}

\section{Introduction}
With a variety of real-world applications, link prediction has been recognized of great importance and attracted increasing attention from the research community in past decade \citep{lu2011link,martinez2017survey}. For instance, link prediction methods could help infer potential protein-protein interactions to efficiently save human effort on blind checking \citep{airoldi2008mixed}. Also, link prediction techniques could be used to predict new friendships between uses on social media, or to discover potential user-to-item relationships on E-commerce sites, such that user experience could be improved \citep{adamic2003friends, koren2009matrix}. 

In the literature, heuristic-based methods are probably the most representative link prediction algorithms. The key idea of most existing heuristic methods is to measure the similarity of two target nodes based on their neighborhood information. The success of these heuristics has demonstrated the importance of neighborhood information of target node-pair. However, heuristic methods often have weak applicability and expressiveness in dealing with different types of networks for its simple-form and hand-crafted information of neighborhood. A previous survey found that all of heuristics methods failed to perform consistently across multiple networks \citep{lu2011link}. The needs of prior knowledge or expensive trial and error are inevitable in choosing appropriate heuristics for different networks. Thanks to effective feature learning ability of neural networks, a series of neural link prediction models \citep{zhang2017weisfeiler,kipf2016variational,wang2020neighborhood,wang2019neighborhood,WANG2021107431} were proposed, of which the generalization ability was successfully improved.

Existing neural link prediction methods pay much attention on designing more expressive neural architectures, while some basic properties of the problem are often neglected. For example, most neural models treat link prediction as a binary classification problem and naturally adopt a cross entropy loss function. However, this learning schema seems not to be suitable for the link prediction problem. First, link classification is extremely imbalanced due to the natural sparsity of most graphs. Although under-sampling could be adopted, there would be information loss during sampling process and what ratio of sampling is hard to decide. Second, most link prediction evaluation protocols do not aim at labeling positive pairs as 1 while negative pairs as 0, but ask for ranking positive pairs higher than negative pairs. Therefore, employing cross-entropy function seems not to be so direct to the objective of the link prediction task.

Based on above understanding and our previous research \citep{wang2020neighborhood,wang2019neighborhood,WANG2021107431}, we provide an effective and generic pairwise learning neural link prediction framework in this paper, named \textbf{PLNLP}. The framework adopts a pairwise learning to rank schema and consists of four main components, i.e., neighborhood encoder, link predictor, negative sampler and objective function. The neighborhood encoder aims at extracting expressive neighborhood information of input node-pair. Any generic graph neural convolution, such as GCN \citep{kipf2016semi} and SAGE \citep{hamilton2017inductive}, or link prediction specific neural architecture, such as SEAL \citep{zhang2018link}, NANs \citep{wang2020neighborhood} and HalpNet \citep{WANG2021107431}, could be employed as neighborhood encoder. For link predictor, we design different scoring functions, which could be selected based on different types of graphs. In negative sampler, we provide several negative sampling strategies, which are problem specific. As for objective function, we propose to use an effective ranking loss, which approximately maximizes the standard ranking metric AUC. We evaluate the proposed PLNLP framework on 4 link property prediction datasets of Open Graph Benchmark (OGB) \citep{hu2020open}, including \texttt{ogbl-ddi}, \texttt{ogbl-collab}, \texttt{ogbl-ppa} and \texttt{ogbl-ciation2}. PLNLP with basic neural architecture achieves \textbf{top 1} performance on \texttt{ogbl-ddi} and \texttt{ogbl-collab}, and \textbf{top 2} performance on \texttt{ogbl-ciation2}. The performance demonstrates the effectiveness of PLNLP.

\section{Related Work}
Existing link prediction approaches can be categorized into three families: heuristic feature based, latent embedding based and neural network based.

\textbf{Heuristic Methods}: Most heuristics measure node similarity with neighborhood information. Popular heuristics include first-order methods common neighbors, Jaccard index \citep{salton1986introduction} and preferential attachment \citep{liben2007link}; second-order methods, i.e., Adamic-Adar \citep{adamic2003friends}, resource allocation \citep{zhou2009predicting}; and high-order heuristic SimRank \citep{jeh2002simrank}. These heuristics often fail to capture complex latent formation features.

\textbf{Embedding-based Methods}: Embedding based methods aim at learning latent node features. The most classical one is matrix factorization (MF) method \citep{menon2011link}, which aims at reconstructing adjacency matrix.  Besides, a series of unsupervised network representation learning models \citep{perozzi2014deepwalk,tang2015line,grover2016node2vec,hamilton2017inductive}, are also applicable for link prediction. These methods learn generic latent embeddings by preserving structure proximities from a probabilistic view and predict links by composing node embeddings as edge features. PNRL \citep{wang2017predictive} is a state-of-the-art link prediction specific embedding method, which simultaneously preserves proximities of observed structure and infers hidden links.

\textbf{NN-based Methods}: Recently, some neural network-based link prediction models were developed, which explore non-linear deep structural features with neural layers. Variational graph auto-encoders \citep{kipf2016variational} predict links by encoding graph with graph convolutional layer \citep{kipf2016semi}. Another two state-of-the-art neural models WLNM \citep{zhang2017weisfeiler} and SEAL \citep{zhang2018link} use graph labeling algorithm to transfer union neighborhood of two nodes (enclosing subgraph) as meaningful matrix and employ convolutional neural layer or a novel graph neural layer DGCNN \citep{zhang2018end} for encoding. 

Besides, in our previous work, we proposed a series of neighborhood attention neural networks \citep{wang2020neighborhood,wang2019neighborhood,WANG2021107431}, in which different attention mechanisms were designed to encode neighborhood information specific for link prediction problem. For instance, in \citep{wang2020neighborhood,wang2019neighborhood}, we proposed cross neighborhood attention and interactive attention mechanisms to capture structural interactions between neighborhoods of the target node-pair.

\section{Preliminaries}
\subsection{Graphs}
Generally, a graph (network) is represented as $G=(V,E)$, where $V=\{v_1, ..., v_N\}$ is the set of nodes, $E\subseteq  V\times V$ is the set of links, and the total number of distinct nodes is $N$. Also, a graph is often denoted as an adjacency matrix $\mathbf{A}$, where $A_{i,j} = 1$ if there is a link from node $v_i$ to $v_j$, otherwise $A_{i,j} = 0$. $\mathbf{A}$ will be symmetric, if the graph is undirected. 

\subsection{Neighborhood of Node}
We use $\mathcal{N}^h(v_i)$ to represent the $h$-hop neighborhood of node $v_i\in V$, which is the set of nodes whose distance to $v_i$ (represented as $d(v_i, v_j)$) is not greater than $h$. In this paper, we focus on unweighted graph, thus the distance function $d(v_i, v_j)$ is directly computed as the length of the shortest path between $v_i$ and $v_j$. We call $v_i$ the center node and $v_j \in \mathcal{N}^h(v_i)$ the neighboring node within $h$-hop. To make the neighborhood also include the unique information of the center node, we define that the center node $v_i$ is a neighboring node of itself, such that $v_i \in \mathcal{N}^h(v_i)$.

\subsection{Neighborhood Subgraph of Node-Pair}
We use $\mathcal{G}^h(v_i, v_j)$ to represent the $h$-hop neighborhood subgraph of the node pair $(v_i, v_j)$, which is extracted from the whole graph $\mathcal{G}$. Formally, for any node $v_k$ in the neighborhood subgraph $\mathcal{G}^h(v_i, v_j)$, it should satisfy $d(v_k, v_i)\leq h$ and $d(v_k, v_j)\leq h$, i.e., $v_k \in \mathcal{N}^h(v_i)\cup \mathcal{N}^h(v_j)$.

\subsection{Link Prediction} Link prediction problems are categorized as temporal link prediction which predicts potential new links on an evolving network, and structural link prediction which infers missing links on a static network. In this paper, we focus on \textit{structural link prediction}. Given the partially observed structure of a network, the goal of it is to predict the unobserved links. Formally, given a partially observed network $G=(V,E)$, we represent the set of node-pairs with unknown link status as $E^?=V\times V-E$, then the goal of structural link prediction is to infer link status of node-pairs in $E^?$.

\section{PLNLP Framework}
\begin{figure*}[tbp]
\centering
\includegraphics[width=15cm]{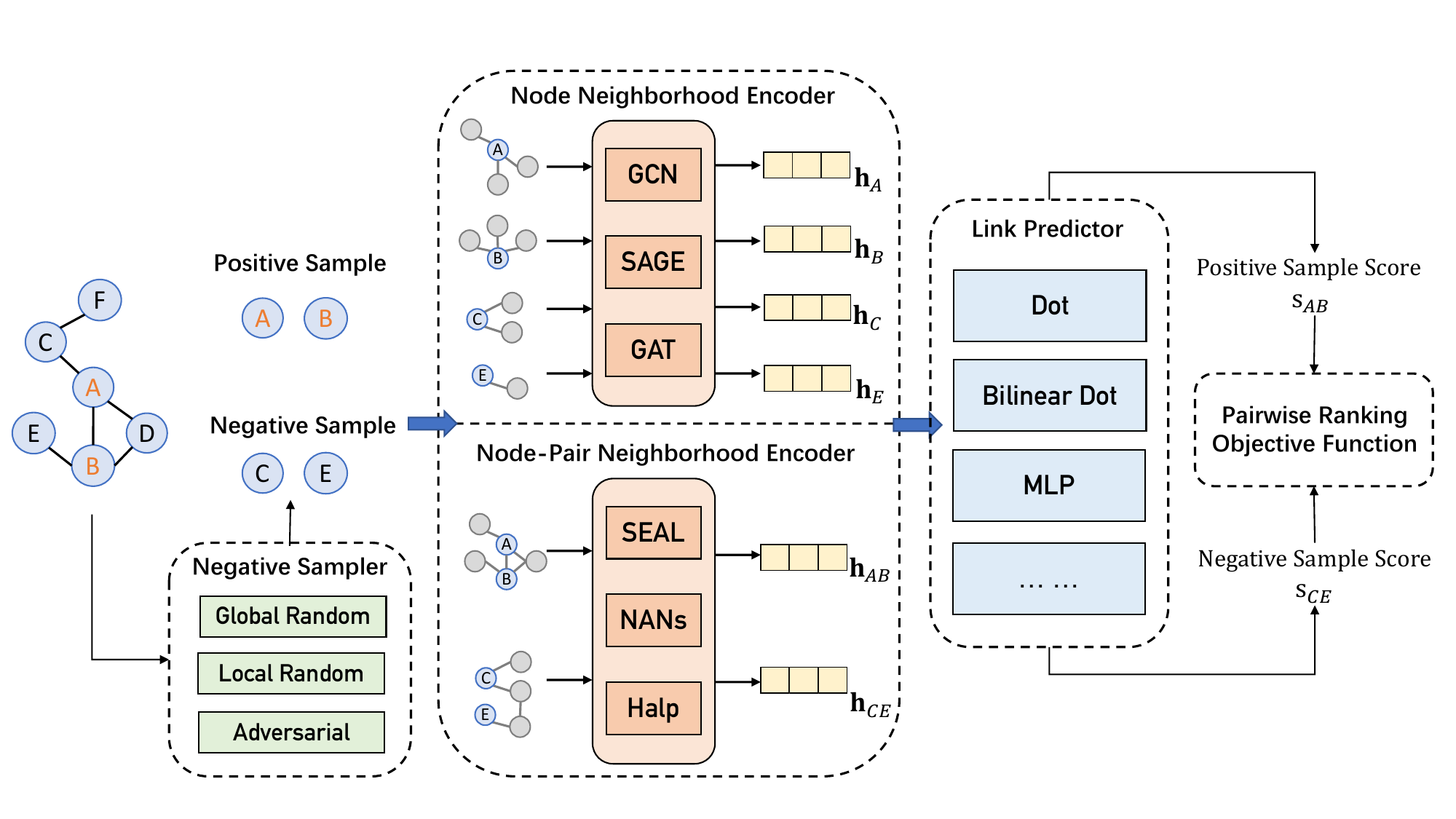}
\caption{PLNLP Framework}
\label{fig:model}
\end{figure*}

The proposed framework is illustrated as Figure \ref{fig:model}. Given an input graph, negative sampler aims to draw negative samples and form training pairs. A training pair consists of a positive sample, which is a node-pair with an observed edge in input graph, and a negative sample, which is a node-pair drawn by negative sampler. Neighborhood encoder is used to extract neighborhood information of both positive and negative samples as the hidden representations. Given the hidden representations, link predictor will calculate link scores of both samples. With link scores of training pairs, the model parameters will be optimized based on the pairwise ranking objective function.

\subsection{Neighborhood Encoder}
Neighborhood information has proved crucial for link prediction. Therefore, we propose to use neighborhood neural encoder to extract structural information of input samples. We consider two kinds of neighborhood encoder in this paper. One is Node Neighborhood Encoder (NNE), which encodes the two nodes of a input sample with their own neighborhood as two hidden representations, separately. Any generic graph neural networks (GNN), e.g, GCN, GraphSAGE and GAT, could be employed as NNE. Assume that input sample is $(v_i, v_j)$, NNEs aim to extract hidden representations of the input sample as follows:
\begin{equation}
 \mathbf{h}_{i} = \textup{NNE}(\mathbf{x}_i, \{\textbf{x}_k\ | {v_k \in \mathcal{N}^h(v_i)} \}), \ \ \ \ 
  \mathbf{h}_{j} = \textup{NNE}(\mathbf{x}_j, \{\textbf{x}_l\ | {v_l \in \mathcal{N}^h(v_j)} \})
\end{equation}
where $\mathbf{x}_i$ generally represents the input feature of node $v_i$. If there is no input features, $\mathbf{x}_i$ represents a embedding vector of node $v_i$, which is trainable parameter. $\mathbf{x}_i$ could also represent the concatenation of input feature and node embedding. In this framework, we only consider homogeneous graph, which means that all nodes share a same NNE.

The other kind of neighborhood encoder is Edge level Neighborhood Encoder (ENE), or called node-pair neighborhood encoder. Recently, a series of ENEs, such as SEAL, NIAN and HalpNet, were proposed specifically for link prediction problem. The main advantage of ENEs is capturing structural interactions between the neighborhoods, which are ignored in NENs. ENEs often consider the neighborhood subgraph of a sample (node-pair) as input, and encode it as one hidden representation. Assume that input sample is $(v_i, v_j)$, ENEs derive a hidden representation of the input sample as follows:
\begin{equation}
 \mathbf{h}_{ij} = \textup{ENE}(\mathbf{x}_i, \mathbf{x}_j,\{\textbf{x}_k\ | {v_k \in \mathcal{G}^h(v_i, v_j)} \})
\end{equation}
Similarly, $\mathbf{x}_i$ represents the input feature, or trainable embedding, or the concatenation of input feature and embedding. 

\subsection{Link Score Predictor}
After deriving the hidden representations either in node level or node-pair (edge) level, the framework will calculate a linking score of the input sample. We provide several selections of the scoring function.

\textbf{Dot Predictor}. If we use NNE to derive $\mathbf{h}_i$ and $\mathbf{h}_j$ of the input sample $(v_i, v_j)$, we can simply use a dot operator to derive the score:
\begin{equation}
 s_{ij} = \mathbf{h}_i \cdot \mathbf{h}_j
\end{equation}

\textbf{Bilinear Dot Predictor}. Dot operator can be only used for undirected graph due to its commutative property. For directed graph, we can adopt bilinear dot operator to make the scoring function not commutative:
\begin{equation}
 s_{ij} = \mathbf{h}_i \mathbf{W} \mathbf{h}_j
\end{equation}
where $\mathbf{W}$ is a learnable matrix.

\textbf{MLP Predictor}. We can also employ a multi-layer perceptron (MLP) as the link predictor. If we use ENEs to obtain hidden representation of the input sample, the predictor is as follow:
\begin{equation}
 s_{ij} = \textup{MLP}(\mathbf{h}_{ij} )
\end{equation}
If we use NNEs to obtain hidden representations, there are several possible forms of MLP's input. If the graph is undirected, we can adopt a widely used commutative operator, i.e, hadamard product $\odot$:
\begin{equation}
 s_{ij} = \textup{MLP}(\mathbf{h}_{i}\odot \mathbf{h}_{j})
\end{equation}
If the graph is directed, we would prefer a non-commutative operator, such as concatenation $||$:
\begin{equation}
 s_{ij} = \textup{MLP}(\mathbf{h}_{i} || \mathbf{h}_{j})
\end{equation}

\subsection{Pairwise Learning with Ranking Objective}
Due to the sparsity of networks, there often exists extreme imbalance between linked pairs and non-linked pairs. Meanwhile, most link prediction tasks do not aim at labeling positive pairs as 1 while negative pairs as 0, but ask for ranking positive pairs higher than negative pairs. To be consistent with the general objective of link prediction, we adopt the ranking idea for model learning, which can be formalized as:
\begin{equation}
 s_{ij} > s_{kl},\forall{(v_i, v_j)\in E}\   \textup{and}\   \forall{(v_k, v_l)\in E^-}
\end{equation}
where $s_{ij}$ and $s_{kl}$ are the output scores of link predictor, $E^-$ is the set of true non-linked pairs. In fact, the above learning objective is equivalent to maximize the Area Under the Curve (AUC), which is interpreted as the probability of a positive sample ranking higher
than a negative sample. The empirical AUC value is defined as follow:
\begin{equation}
 \textup{AUC} = \sum_{(v_i, v_j)\in E} \sum_{(v_k, v_l) \in E^-}\frac{\mathbbm{1}[f_\theta(v_i, v_j) > f_\theta(v_k, v_l)]}{|V\times V|}
\end{equation}
where $\mathbbm{1}[\cdot]$ is an indicator function that equals to 1 if $f_\theta(v_i, v_j) > f_\theta(v_k, v_l)$, otherwise equals to 0. $f_\theta(v_i, v_j) = s_{ij}$ represents the output of the neural link prediction model, where $\theta$ denotes all parameters of the model. Optimizing AUC is not straightforward since the gradient of this function is either zero or not defined. 
Various techniques have been proposed to approximate the AUC with a surrogate function. There are several possible selections of surrogate functions, such as pairwise hinge loss, logistic loss or exponential loss. In this paper, we simply select the squared least surrogate loss, which is proved consistent with AUC theoretically \citep{gao2015consistency}. Our framework is flexible to adopt any other surrogate function that approximates AUC. The base AUC-optimization objective function is defined as follow:
\begin{equation}
\label{eq:oriloss}
O_{\textup{\scriptsize{AUC}}} = \min_{\theta}\sum_{{(v_i, v_j)\in E, (v_i, v_k)\in E^-}}\left(1 -  f_{\theta}\left(v_i, v_j\right) + f_{\theta}\left(v_i, v_k\right)  \right)^2 + \frac{\lambda}{2}||\theta||^2
\end{equation}
The above function forces the margin between positive samples and negative samples to be 1. In some situations, this constraint is too strict for the optimization. It can be relaxed by combining the squared hinge loss with above function:
\begin{equation}
\label{eq:hingeloss}
O_{\textup{\scriptsize{Hinge-AUC}}} = \min_{\theta}\sum_{{(v_i, v_j)\in E, (v_i, v_k)\in E^-}}\left(\max\left(0, 1 -  f_{\theta}(v_i, v_j) + f_{\theta}(v_i, v_k)  \right)\right)^2 + \frac{\lambda}{2}||\theta||^2
\end{equation}
The above function only forces the margin between positive samples and negative samples to be larger than 1. 

Furthermore, the margin may not be fixed as 1 if weights on training edges (positive sample) are expected to be modeled. A straightforward way of introducing sample weights is as follows:
\begin{equation}
\label{eq:weighthingeloss}
O_{\textup{\scriptsize{Weighted-Hinge-AUC}}} = \min_{\theta}\sum_{{(v_i, v_j)\in E, (v_i, v_k)\in E^-}}\gamma_{ij} \left(\max\left(0, \gamma_{ij} -  f_{\theta}(v_i, v_j) + f_{\theta}(v_i, v_k)  \right)\right)^2 + \frac{\lambda}{2}||\theta||^2
\end{equation}
where $\gamma$ is an adaptive margin, which may correspond to normalized weights of training edges.

In above objective functions, to prevent over-fitting problem, we use the L2 regularization on parameters with a weight $\lambda$. Given a positive pair $(v_i, v_j)$ and a sampled negative pair $(v_k, v_l)$, the parameters $\theta$ of the model are optimized by the stochastic gradient descent (SGD) method. For most cases, we use the basic objective function in Eq. \ref{eq:oriloss}. When sample weights are considered, the objective function of Eq. \ref{eq:weighthingeloss} is used.

\subsection{Negative Sampling}
In practice, true non-linked set $E^-$ is not available in the training data. A conventional strategy is randomly sampling a negative node-pair $(v_k, v_j)$, which has unknown link status and is assumed as negative samples. For different problems or types of graphs, we may have different sampling strategies. 

\textbf{Global Sampling}. Global sampling represents that, for each positive sample, we uniformly sample a negative node-pair from the set $E^?=V\times V-E$. This strategy is suitable for the problem seeking for global ranking performance. For example, in the protein-protein interaction, we are interested in potential node-pairs, which are worth performing further analysis on, among all possible node-pairs.

\textbf{Local Sampling}. Local sampling represents that, for a positive sample $(v_i, v_j)$, we firstly select an anchor node saying $v_i$, then uniformly sample a node $v_k$ and regard $(v_i, v_k)$ as the negative sample. Instead of uniform distribution, other distribution, e.g, the power of node degrees, can be applied to sample the negative node $v_k$. This strategy is appropriate to the situation that aims to obtain good ranking for individual nodes. For example, in a recommendation system, we would like to recommend a good ranking list of items to each individual user.

\textbf{Adversarial Sampling}. The performance of random sampling strategy is not always stable due to complete randomness. Similar problems of random negative sampling have also been found in other tasks, e.g., knowledge graph embedding \citep{Wang2018IncoGAN,cai2017kbgan} and image retrieval \citep{wu2017sampling}. In our previous work \citep{wang2020neighborhood}, we proposed to use adversarial learning technique to generate negative samples instead of random sampling. We designed a generative model to generate high quality negative samples, which aims at making difficulties to link prediction model. In this way, link prediction model and negative sample generator play an adversarial game. By continuously providing high quality negative samples, adversarial sampling more robust than random sampling. We leave the evaluation of adversarial sampling on ogb datasets as future work.

\begin{figure*}[tbp]
\centering
\includegraphics[width=15cm]{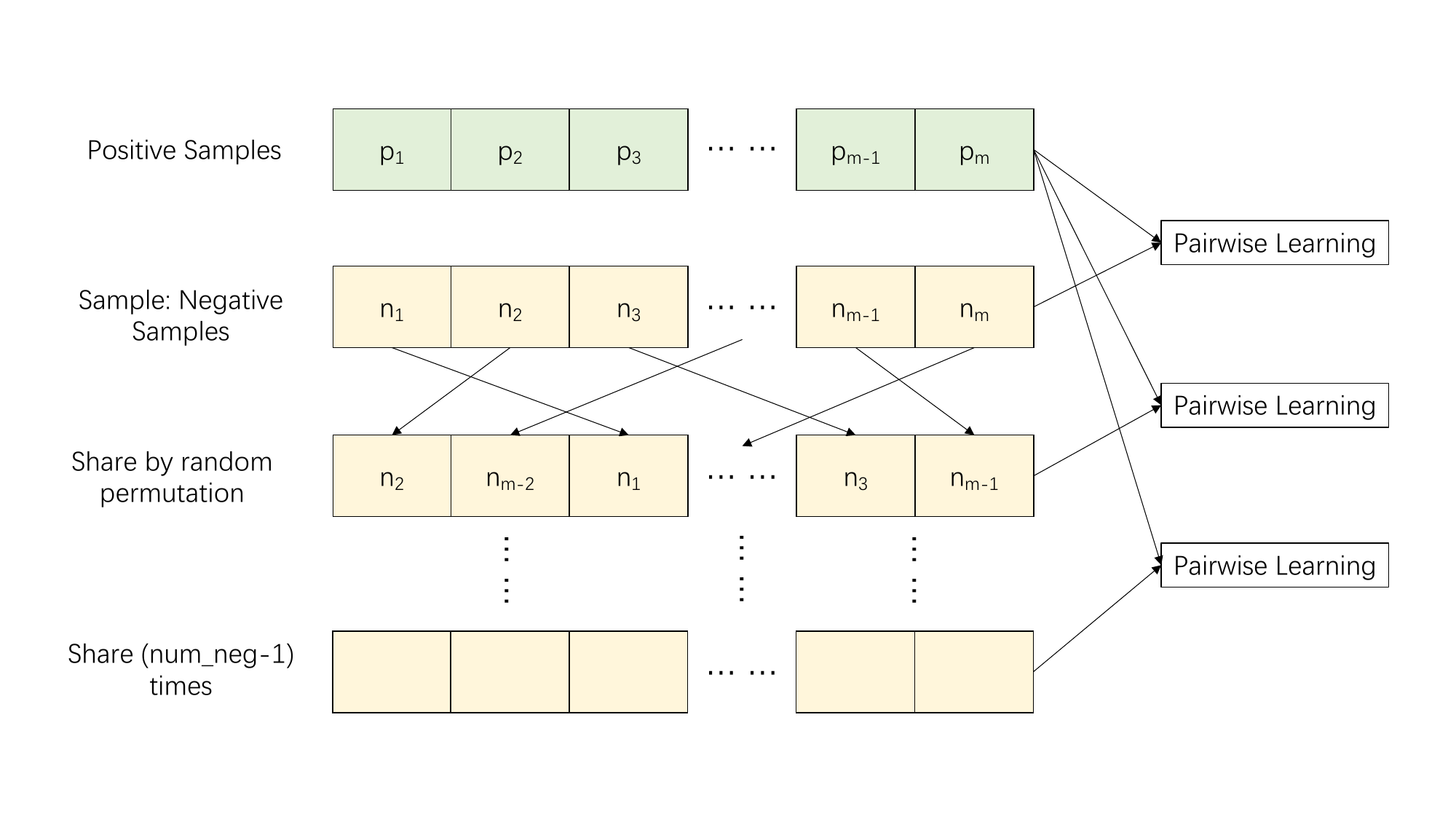}
\caption{Negative Sample Sharing Mechanism}
\label{fig:neg}
\end{figure*}

\textbf{Negative Sample Sharing}. Since the framework adopts pairwise schema, each negative sample can only be used for one positive sample once, which is not efficient. To make better use of negative samples, we propose a negative sample sharing mechanism. As shown in Figure \ref{fig:neg}, assume that the total number of positive sample is $m$, we firstly draw $m$ negative samples and construct $m$ training pairs with same indexes. Given the negative samples, the sharing mechanism will random permute the indexes of negative samples, and form $m$ new training pairs. The hyper-parameter \texttt{num\_neg} indicates the mechanism will random permute \texttt{(num\_neg-1)} times of negative samples. By using this sharing mechanism, we could create $m\times\texttt{num\_neg}$ training pairs by only sampling $m$ negative samples at each training epoch.

\subsection{Data Augmentation with Random Walk}
In some graphs, high-order structure information play a important role. Although increasing the number of GNN layers could model the high-order information, it also may leads to over-smoothing problem and low efficiency. To this end, we propose to use data augmentation to introduce high-order information at the input. A general technique to sample high order information is the Random Walk. Given all nodes in the graph, we use the basic random walk method to sample the high-order pairs. Assume the start point node is $v_i$ and its random walk is $\textup{RW}(v_i)= \{v_{k+1}, ..., v_{k+l}\}$, where $l$ represents the walk length , then the set of positive samples is augmented as : 
\begin{equation}
E_{\textup{aug}} = E \cup \{(v_i, v_j) |  v_j \in \textup{RW}(v_i), \forall v_i \in V\}
\end{equation}
Meanwhile, the augmented pairs are associated with weights based on the steps of walks. For example, in the walk, $\textup{RW}(v_i)= \{v_{k+1}, ..., v_{k+l}\}$, the augmented pair $(v_i, v_{k+l})$ is associated with the weight $1/l$. With different weights of augmented pairs in $E_{\textup{aug}}$, we find that using the weight-adaptive objective function in Eq. \ref{eq:weighthingeloss} is more effective. Therefore, it is suggested using this objective function when random walk augmentation is adopted.

\section{Evaluation on OGB}
Our code for evaluation is available at \texttt{https://github.com/zhitao-wang/PLNLP}.
\subsection{Datasets and Evaluation Metrics}
We evaluate the link prediction ability of PLNLP on Open Graph Benchmark (OGB) data \citep{hu2020open}. Four data sets with different graph types are evaluated, including ogbl-ddi, ogbl-collab, ogbl-citation2 and ogbl-ppa.

\textbf{ogbl-ddi}: The dataset is a homogeneous, unweighted, undirected graph, representing the drug-drug interaction network. Each node represents a drug. Edges represent interactions between drugs. 

The task is to predict drug-drug interactions given information on already known drug-drug interactions. The performance is evaluated by Hits@20: each true drug interaction is ranked among a set of approximately 100,000 randomly-sampled negative drug interactions, and count the ratio of positive edges that are ranked at 20-place or above.

\textbf{ogbl-collab}: The dataset is an undirected graph, representing a subset of the collaboration network between authors indexed by MAG. Each node represents an author and edges indicate the collaboration between authors. All nodes come with 128-dimensional features. 

The task is to predict the future author collaboration relationships given the past collaborations. Evaluation metric is Hits@50, where each true collaboration is ranked among a set of 100,000 randomly-sampled negative collaborations.

\textbf{ogbl-ppa}: The dataset is an undirected, unweighted graph. Nodes represent proteins from 58 different species, and edges indicate biologically meaningful associations between proteins. 

The task is to predict new association edges given the training edges. Evaluation metric is Hits@100, where each positive edge is ranked among 3,000,000 randomly-sampled negative edges.

\textbf{ogbl-citation2}: The dataset is a directed graph, representing the citation network between a subset of papers extracted from MAG. Each node is a paper with 128-dimensional word2vec features. 

The task is to predict missing citations given existing citations. The evaluation metric is Mean Reciprocal Rank (MRR), where the reciprocal rank of the true reference among 1,000 negative candidates is calculated for each source paper, and then the average is taken over all source papers.

\subsection{Evaluation Settings and Results}
The detailed settings of PLNLP in this paper are shown in Table \ref{table:settings}. We only employ basic node level neighborhood encoder, e.g., GCN or SAGE, to demonstrate the effectiveness of the proposed framework. Some well-design edge level neighborhood encoder specific for link prediction, such as SEAL, NANs and HalpNet, may further improve the performance. But due to low efficiency of edge level neighborhood encoders, we leave this part in the future work. We treat all training datasets as unweighted and undirected graphs. As for MLP predictor, we use hadamard product to get the input of MLP. It is worth noting that we use validation set for training on \texttt{ogbl-collab}, which is allowed by OGB. Meanwhile, we employ the trick from HOP-REC that we only use training edges after year 2010 in \texttt{ogbl-collab}. Furthermore, we use random walk augmentation with a walk length 10 for \texttt{ogbl-collab}.
\begin{table}[tp]
\caption{Settings of PLNLP on OGB Datasets}
  \label{table:settings}
  \renewcommand\arraystretch{1.5}
  \begin{tabular}{|c|c|c|c|c|c|c|} \hline
     & Loss Func. & Encoder & Predictor & Neg. Sampler  & \makecell[c]{Other \\ Parameters}\\ \hline
    \texttt{ddi} &  $O_{\textup{\scriptsize{AUC}}}$ & \makecell[c]{SAGE \\ layer = 2 \\ dim = 512 \\ dropout = 0.3} & \makecell[c]{MLP \\ layer = 2 \\ dim = 512 \\ dropout = 0.3} & \makecell[c]{GLOBAL \\ num\_neg = 3 } & \makecell[c]{node emb. = 512 \\ node feat. = no \\ lr = 0.001 \\ epochs = 500 } \\
    \hline
    \texttt{collab} & $O_{\textup{\scriptsize{Weighted-Hinge-AUC}}}$ & \makecell[c]{SAGE \\ layer = 1 \\ dim = 256 \\ dropout = 0.3} & DOT & \makecell[c]{GLOBAL \\ num\_neg = 1 } & \makecell[c]{node emb. = 256 \\ node feat. = no \\ lr = 0.001 \\ epochs = 800 \\ random walk aug. = yes \\ walk length = 10} \\
    \hline
    \texttt{ppa} & $O_{\textup{\scriptsize{AUC}}}$ & \makecell[c]{SAGE \\ layer = 2 \\ dim = 256 \\ dropout = 0.3} & DOT & \makecell[c]{GLOBAL \\ num\_neg = 3 } & \makecell[c]{node emb. = 256 \\ node feat. = yes \\ lr = 0.001 \\ epochs = 200 } \\
    \hline
    \texttt{citation2} & $O_{\textup{\scriptsize{AUC}}}$ & \makecell[c]{GCN \\ layer = 2 \\ dim = 200 \\ dropout = 0.0} & \makecell[c]{MLP \\ layer = 2 \\ dim = 200 \\ dropout = 0.0} & \makecell[c]{LOCAL \\ num\_neg = 3 }  & \makecell[c]{node emb. = 50 \\ node feat. = yes \\ lr = 0.001 \\ epochs = 100 } \\

    \hline
    \end{tabular}
\end{table}

Following OGB rules, we evaluate PLNLP with 10 runs, without fixing random seed. As for other state-of-the-art methods, we just copy the results from OGB official leader board.  

\begin{table}[tp]
\centering
\caption{Link Prediction Performance on OGB (Test Performance)}
  \label{table:lp_performance}
  \renewcommand\arraystretch{1.1}
  \begin{tabular}{c|cccc} \hline
     & \makecell[c]{\texttt{ogbl-ddi}\\{Hits@20(\%)}} & \makecell[c]{\texttt{ogbl-collab}\\{Hits@50(\%)}} & \makecell[c]{\texttt{ogbl-ppa}\\{Hits@100(\%)}} & \makecell[c]{\texttt{ogbl-citation2}\\{MRR(\%)}} \\ \hline
    \textbf{CN} & $17.73\pm0.00$ & $61.37\pm0.00$ & $27.65\pm0.00$ & $51.47\pm0.00$  \\
    \textbf{AA} & $18.61\pm0.00$ & $64.17\pm0.00$ & $32.45\pm0.00$ & $51.89\pm0.00$  \\
    \textbf{RA} & $-$ & $-$ & $49.33\pm0.00$ & $-$  \\
    \textbf{AA+Proposal Set} & $-$ & $65.48\pm0.00$ & $-$ & $-$ \\
    \textbf{RA+Proposal Set} & $-$ & $-$ & $\textbf{53.24}\pm\textbf{0.00}$ & $-$ \\
    \hline
    \textbf{MF} & $13.68\pm4.75$ & $38.86\pm0.29$ & $32.29\pm0.94$ & $51.86\pm4.43$ \\
    \textbf{DeepWalk} & $22.46\pm2.90$& $50.37\pm0.34$ & $23.02\pm1.63$ & $-$ \\
    \textbf{Node2vec} & $23.26\pm2.09$ & $48.88\pm0.54$ & $22.26\pm0.83$ & $-$ \\
    \textbf{HOP-REC} & $-$ & $70.12\pm0.16$ & $-$ & $-$  \\
    \hline
    \textbf{SAGE} & $53.90\pm4.74$ & $54.63\pm1.12$ & $16.55\pm2.40$ & $82.60\pm0.36$ \\
    \textbf{GCN} & $37.07\pm5.07$ & $47.14\pm1.45$ & $18.67\pm1.32$ & $84.74\pm0.31$  \\
    \textbf{SEAL} & $30.56\pm3.86$ & $64.74\pm0.33$ & $48.80\pm3.16$ & $\textbf{87.67}\pm\textbf{0.32}$  \\
    \textbf{SAGE+Proposal Set} & $74.95\pm3.17$ & $-$ & $-$ & $-$ \\
    \textbf{CFLP (w/ JKNet)} & $86.08\pm1.98$ & $-$ & $-$  & $-$\\
    \textbf{SAGE+Edge Attr} & $87.81\pm4.47$ & $-$ & $-$  & $-$\\
    
    \hline
    \textbf{PLNLP} & $\textbf{90.88}\pm\textbf{3.13}$ & $\textbf{70.59}\pm\textbf{0.29}$ & $32.38\pm2.58$ & $\textbf{84.92}\pm\textbf{0.29}$  \\
    \hline
    \end{tabular}
\end{table}

The averaged results with standard deviation are reported in the Table \ref{table:lp_performance}. Only with basic graph neural architectures, PLNLP achieves top 1 performance on \texttt{ogbl-ddi} and \texttt{ogbl-collab}, and top 2 performance on \texttt{ogbl-ciation2}. This significantly demonstrates the effectiveness of PLNLP.

\subsection{Ablation Study}
Furthermore, we compare PLNLP against the generic classification learning framework where the loss function is cross-entropy. In this ablation study, we keep same neural architecture (same encoder, predictor with same parameters as reported in Table \ref{table:settings}) in the two frameworks. We use basic AUC objective function Eq.\ref{eq:oriloss} for all datasets and do not use random walk augmentation for \texttt{ogbl-collab} in this study. To guarantee fairness, we use same negative sampling strategies and use the same number of negative samples at each epoch. The results are shown in Table \ref{table:ablation}. It is found that PLNLP remarkably outperforms the generic classification learning schema, which indicates the proposed pairwise learning could maximize the performance of graph neural models on link prediction problem.

\begin{table}[tp]
\centering
\caption{PLNLP vs. Classification Schema}
  \label{table:ablation}
  \renewcommand\arraystretch{1.3}
  \begin{tabular}{c|cccccccc} \hline
     & \multicolumn{2}{c}{\makecell[c]{\texttt{ogbl-ddi}\\{{Hits@20(\%)}}}}  & \multicolumn{2}{c}{\makecell[c]{\texttt{ogbl-collab}\\{{Hits@50(\%)}}}} & \multicolumn{2}{c}{\makecell[c]{\texttt{ogbl-ppa}\\{{Hits@100(\%)}}}} & \multicolumn{2}{c}{\makecell[c]{\texttt{ogbl-citation2}\\{{MRR(\%)}}}} \\ 
     \cline{2-9}
     & Test & Valid & Test & Valid & Test & Valid & Test & Valid  \\
     \hline
    \textbf{Classification} & 70.70 & 68.02 & 64.97 & 99.17 & 16.55 & 17.24 & 84.64 & 84.73  \\
    \hline
    \textbf{PLNLP} & \textbf{90.88} & \textbf{82.42} & \textbf{68.72} & \textbf{100.00} & \textbf{32.38} & \textbf{31.62} & \textbf{84.92} & \textbf{84.90} \\
    \hline
    \end{tabular}
\end{table}

\subsubsection*{Acknowledgments}
The authors greatly thank the great support for advanced research from departments of WeChat Pay and WeChat Search.

\bibliographystyle{plainnat}
\bibliography{sample}

\end{document}